\documentclass{article}

\usepackage[position, final]{neurips_2026}

\usepackage[utf8]{inputenc} 
\usepackage[T1]{fontenc}    
\usepackage{hyperref}       
\usepackage{url}            
\usepackage{booktabs}       
\usepackage{amsfonts}       
\usepackage{nicefrac}       
\usepackage{microtype}      
\usepackage{xcolor}         

\usepackage{enumitem}

\usepackage{amsmath}

\usepackage{tikz}
\usepackage{tikz-cd}
\usetikzlibrary{arrows.meta, positioning}
\usepackage{wrapfig}

\usepackage{amsmath}
\usepackage{amssymb}
\usepackage{mathtools}
\usepackage{amsthm}

\usepackage{enumitem}
\usepackage{dsfont}

\usepackage[most]{tcolorbox}
\usepackage{xcolor}
\usepackage[dvipsnames]{xcolor}

\theoremstyle{plain}
\newtheorem{theorem}{Theorem}[section]
\newtheorem{proposition}[theorem]{Proposition}

\theoremstyle{definition}

\theoremstyle{remark}

\newcommand{\bv}{\mathbf{v}}
\newcommand{\bU}{\mathbf{U}}

\newcommand{\bV}{\mathbf{V}}

\newcommand{\bZ}{\mathbf{Z}}

\hypersetup{
  colorlinks=true,
  linkcolor=RubineRed,
  citecolor=MidnightBlue,
  urlcolor=MidnightBlue,
  pdfborder={0 0 0}
}

\definecolor{PositionBlue}{HTML}{1F4E79}
\definecolor{PositionBlueLight}{HTML}{F3F7FB}
\definecolor{PositionBlueBorder}{HTML}{B7CCE2}

\newtcolorbox{positionbox}{
  enhanced,
  breakable,
  colback=PositionBlueLight,
  colframe=PositionBlueBorder,
  coltitle=white,
  colbacktitle=PositionBlue,
  fonttitle=\bfseries\sffamily,
  title=Position Statement,
  boxrule=0.7pt,
  arc=2.5mm,
  left=2.5mm,
  right=2.5mm,
  top=1.5mm,
  bottom=1.5mm,
  titlerule=0pt,
  toptitle=1mm,
  bottomtitle=1mm,
  attach boxed title to top left={
    xshift=2.5mm,
    yshift*=-2mm
  },
  boxed title style={
    colback=PositionBlue,
    colframe=PositionBlue,
    arc=2mm,
    boxrule=0pt,
    left=1.8mm,
    right=1.8mm,
    top=0.7mm,
    bottom=0.7mm
  },
  drop fuzzy shadow
}

\title{The Dead Salmons of AI Interpretability}

\author{
Maxime Méloux$^1$ \quad Giada Dirupo$^2$ \quad François Portet$^1$ \quad Maxime Peyrard$^1$ \\
$^1$Université Grenoble Alpes, CNRS, Grenoble INP, LIG \\  $^2$Icahn School of Medicine at Mount Sinai\\
\texttt{\{melouxm, portetf, peyrardm\}@univ-grenoble-alpes.fr}\\
\texttt{giada.dirupo@mssm.edu}\\
}

\begin{document}

\usetikzlibrary{calc}

\maketitle

\begin{abstract}
  In a striking neuroscience study, the authors placed a dead salmon in an MRI scanner and showed it images of humans in social situations. Astonishingly, standard analyses of the time reported brain regions \textit{predictive} of social emotions. The explanation, of course, was not supernatural cognition but a cautionary tale about misapplied statistical inference. In AI interpretability, reports of similar ``dead salmon'' artifacts abound: feature attribution, probing, sparse auto-encoding, and even causal analyses can produce plausible-looking explanations for randomly initialized neural networks. Upon inspection of this phenomenon our position is to argue for a pragmatic \textbf{statistical--causal reframing}: explanations of computational systems should be treated as parameters of a (statistical) model, inferred from computational traces. This perspective goes beyond simply measuring statistical variability of explanations due to finite sampling of input data; interpretability methods become statistical estimators, and findings should be tested against explicit and meaningful \textbf{alternative computational hypotheses}, with uncertainty quantified with respect to the postulated statistical model. It also highlights important theoretical issues, such as the identifiability of common interpretability queries, which we argue is critical to understand the field’s susceptibility to false discoveries, poor generalizability, and high variance.
\end{abstract}

\begin{positionbox}
Interpretability claims about AI systems should be treated as statistical--causal inference claims. Explanations are inferred from finite computational traces under specific assumptions and should therefore be accompanied by uncertainty quantification, meaningful computational null models, robustness checks, and identifiability analyses. The striking false positives revealed by the \textit{dead salmon} findings highlight not only the consequences of neglecting these safeguards, but also a bigger challenge: the prevalence of causally overdetermined structures in neural networks.
\end{positionbox}

\section{Introduction}
\label{sec:introduction}

In 2009, researchers placed a dead salmon in an MRI scanner, showed it photographs of humans in social situations, and ostensibly asked it to judge their emotions \citep{bennett2009neural}. Standard analysis pipelines commonly used at the time surprisingly returned brain voxels as significantly predictive of emotional situations. The error arose from a failure to correct for multiple comparisons within the statistical analysis pipeline. The “dead salmon” demonstration of false positives contributed to a larger reckoning in the field of neuroscience. For instance, an influential study showed that different research groups obtained different results even when analyzing the same dataset and the same research question~\citep{botvinik2020variability}. 
Subsequent work identified several sources of \textit{statistical fragility}. Widely used statistical procedures embedded in standard analysis pipelines were shown to inflate false-positive rates~\citep{eklund2016cluster}, an effect worsened by non-independent analyses producing spuriously large brain–behavior correlations~\citep{vul2009voodoo}. Also, early neuroimaging research was constrained by small samples and limited data availability~\citep{button2013power,marek2022reproducible}, exacerbating overfitting and spurious associations. Moreover, fMRI had been criticized for offering predictive explanations, rather than functional ones, resulting in little clinical relevance~\citep{lyon2017dead}. Finally, reverse inference emerged as a central interpretative problem, given that individual neural systems are not uniquely associated with specific cognitive functions~\citep{poldrack2006can,duncan2000common}. 

\begin{wrapfigure}{r}{0.45\linewidth}
    \centering
    \vspace{-1em}
    \includegraphics[width=\linewidth]{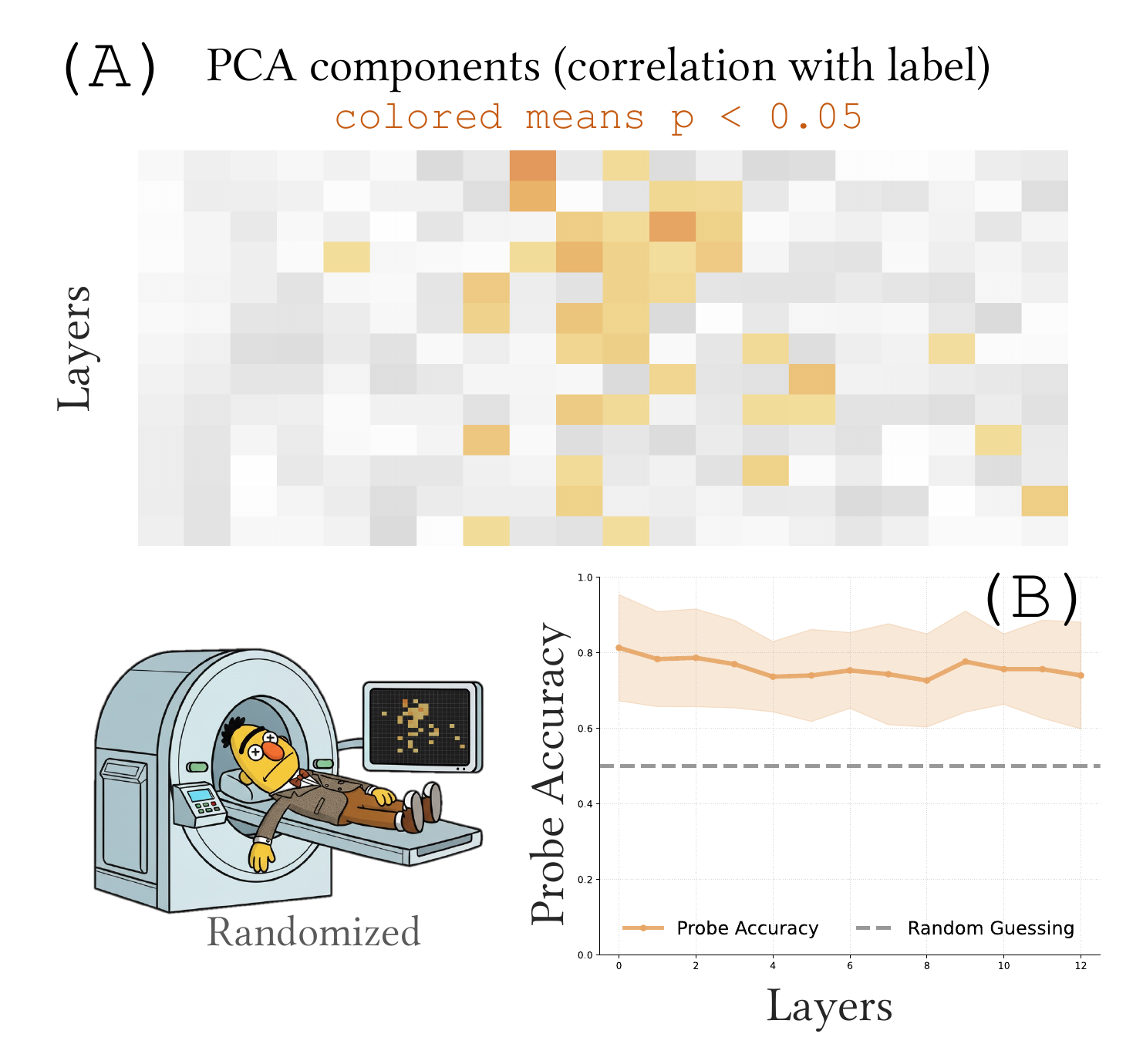}
    \caption{\textbf{Exhibit of minimal dead salmon artifacts}. We extract token representations from a randomly initialized BERT for 300 IMDb sentences and average over sequence length. (A) Each cell is a (layer, principal component) pair of the mean-pooled activations; coloured cells are those whose Pearson correlation with the binary sentiment label has a two-sided $p<0.05$. No correction for multiple comparisons is applied, on purpose: the panel shows what an uncorrected pipeline returns, as in the original study, and these individually weak correlations are not expected to survive a Bonferroni or Benjamini--Hochberg correction~\citep{benjamini1995controlling}. (B) A simple probe trained on layer representations achieves nontrivial cross-validated accuracy.}
    \label{fig:illus}
\end{wrapfigure}

AI interpretability now faces its own \textit{dead salmon} issues, similarly begging for a larger reevaluation of its statistical foundations. A growing body of work has shown that many influential methods, including feature attribution \citep{adebayo2018sanity}, probing classifiers \citep{ravichander-etal-2021-probing}, sparse autoencoders \citep{heap2025sparseautoencodersinterpretrandomly}, circuit discoveries \citep{melouxeverything}, and causal abstractions~\citep{sutter2025nonlinearrepresentationdilemmacausal}, can yield plausible-looking explanations even when applied to \textbf{random neural networks}. In Figure~\ref{fig:illus}, we report a minimal dead salmon artifact from analyzing activations of a fully randomized BERT model in a sentiment analysis task, where both an uncorrected correlation analysis and probing return nominally significant \textit{explanations}.
We use ``dead salmon'' in a narrow sense: running a full analysis pipeline on a system known to contain no relevant mechanism, and observing that it still returns an explanation. Dead salmons are therefore a diagnostic for a subset of \textit{interpretability illusions}~\citep{bolukbasi2021interpretabilityillusionbert,makelov2023subspacelookingforinterpretability}, the broader phenomenon of explanations that look valid without reflecting the computation. The case of Figure~\ref{fig:illus} is the easy one, since a multiple-comparison correction removes it; the harder cases (Appendix~\ref{app:exp}) are those where effects are large relative to chance but not relative to a randomized network. Such striking failure modes are particularly troubling as modern AI systems are increasingly deployed in high-stakes domains where AI interpretability should be essential for transparency, accountability, and error diagnosis~\citep{mehrabi2021survey,barnes2024navigating,ramachandram2025transparentaicaseinterpretability}.
Interpretability methods have the potential surface critical failure modes~\citep{kim2017interpretable,zech2018variable,caruana2015intelligible,meng2022locating,monea2024glitchmatrixlocatingdetecting,nguyen2025deploying} and offer levers for mitigating bias and systematic errors \citep{arrieta2019explainableartificialintelligencexai,kristofik2025bias,lepori-etal-2025-racing}. 

Yet, despite frequent analogies to mature sciences like \textit{neuroscience}~\citep{barrett2019analyzing}, \textit{biology}~\citep{lindsey2025biology}, or \textit{physics} \citep{AL2023-cfg,AllenZhu-icml2024-tutorial} of neural networks, the practice of AI interpretability remains in its early foundational stages. Striking dead-salmon artifacts are accompanied by a general statistical fragility: small perturbations to inputs \citep{Ghorbani_Abid_Zou_2019,Kindermans2019,pmlr-v260-zhang25a} or changes in random initialization \citep{adebayo2018sanity,zafar-etal-2021-lack} can radically change explanations. Explanations often fail to generalize to new settings and input distributions \citep{hoelscherobermaier2023detectingeditfailureslarge}. Also, multiple incompatible explanations can be \textit{discovered} for the same behavior~\citep{melouxeverything,NEURIPS2019_bb836c01}. While the dead salmon study demonstrated a simple statistical oversight correctable through multiple comparison adjustments, AI interpretability's difficulties stem from more fundamental issues. In particular, we argue that, for common interpretability queries, computational traces do not uniquely determine explanations.

    

Beyond neuroscience and AI, such challenges are not unprecedented. Psychology and the social sciences faced a similar reckoning during the replication crisis, when questionable research practices produced widespread false positives \citep{doi:10.1126/science.aac4716,simmons2011false,ioannidis2005most,schimmack2020meta}. These fields responded with methodological reforms: pre-registration, registered reports, increased statistical power, and explicit multiple-comparison corrections \citep{munafo2017manifesto,korbmacher2023replication}. 

Now, AI interpretability can also begin to build its own methodological guardrails. As argued before, this requires both technical innovation and philosophical clarity \citep{miller2019explanation,williams2025mechanisticinterpretabilityneedsphilosophy}. This means clarifying our epistemic goals by answering: what does it mean to ``explain'' a neural network? \citep{lillicrap2019doesmeanunderstandneural,lipton2018mythos} Mechanistic interpretability embodies a type of \textit{scientific realism}, aiming to discover the \textit{one true} explanatory algorithm~\citep{psillos2005scientific,chakravartty2011scientific}. However, there is a significant push-back against the feasibility of this research project~\citep{rudin2019stop,P_ez_2019,saphra2024mechanistic}, motivating a shift toward pragmatic approaches prioritizing the utility of the explanations for specific downstream goals~\citep{zou2025representationengineeringtopdownapproach}. Here, we align with the pragmatic stance~\citep{Dewey1948-DEWRIP,chang2004inventing,potochnik2017idealization}, where explanations are seen as useful models that enable prediction, manipulation, and control~\citep{van1980scientific,cartwright1983laws}. 

\textbf{Our position}, stated above, is supported by four contributions. First, a \emph{diagnosis} (Sections~\ref{sec:description} and~\ref{sec:issue}) that separates the failure modes usually grouped under ``fragility'' and, within non-identifiability, distinguishes \emph{underspecification}, which mainly affects predictive methods (probes, attribution, SAEs), from \emph{overdetermination}, which mainly affects causal methods (ablation, circuit discovery, interchange interventions). Second, a \emph{framework} (Section~\ref{sec:stats_inf_view}) in which explanations are surrogate models fitted for a task $(\mu,\mathcal{E},D)$. Probe accuracy, circuit faithfulness, the hypothesis tests of \citet{shi2024hypothesistestingcircuithypothesis}, and causal abstraction with interchange-intervention accuracy~\citep{geiger2025causalabstractiontheoreticalfoundation} are special cases, so that methods which look very different can be compared through the assumptions they make. Third, a \emph{reporting pipeline} (Section~\ref{sec:rec}): declare the task, report residual risk and uncertainty, and test against explicit computational nulls. Fourth, a \emph{research agenda} (Section~\ref{sec:discussion}): characterizing which tasks are identifiable from which query families. Section~\ref{sec:alternatives} discusses alternative views and settings where the framework does not apply.

\section{The Statistical Fragility of AI Interpretability}
\label{sec:description}
Reports documenting the failure modes of interpretability methods are frequent and share a recurring theme: a general statistical fragility, most strikingly illustrated by dead salmon artifacts. We summarize them here; Appendix~\ref{app:related} gives a fuller overview.

\medskip

\noindent\textbf{Feature attribution.}
Saliency maps can remain visually plausible after model weights are randomized~\citep{adebayo2018sanity}, can be manipulated while predictions are unchanged~\citep{NEURIPS2019_bb836c01}, and are unstable under minor input transformations~\citep{Ghorbani_Abid_Zou_2019}. Impossibility results show that no attribution method satisfies intuitive desiderata across broad model classes~\citep{bilodeau2024impossibility}.

\medskip

\noindent\textbf{Probing, SAEs and concept-based methods.}
Probes recover information with high accuracy from randomized embeddings~\citep{conneau-etal-2018-cram,hewitt-manning-2019-structural} and detect what is encoded rather than what is used~\citep{ravichander-etal-2021-probing}; capacity control~\citep{voita-titov-2020-information,pimentel-etal-2020-information,belinkov-2022-probing} and amnesic probing~\citep{elazar-etal-2021-amnesic} were proposed in response. SAEs find apparently interpretable features in randomly initialized transformers~\citep{heap2025sparseautoencodersinterpretrandomly} and often fail to generalize~\citep{kantamneni2025sparseautoencodersusefulcase}. Concept-based explanations show illusions at the neuron level~\citep{bolukbasi2021interpretabilityillusionbert}, inconsistency, and high variance~\citep{ramaswamy2023overlooked,piratla2024estimationconceptexplanationsuncertainty}. These problems also arise in the interpretability of latent reasoning models, where proposed explanations of how latent thoughts shape models' answers have been called into question by findings that these thoughts are often not causally responsible for the resulting answers~\citep{aswal2026observablepatternsexplanationscausalgeometric,cui2026how,hassija2024interpreting}

\medskip

\noindent\textbf{Causal approaches and mechanistic interpretability.}
Causal mediation and activation patching~\citep{NEURIPS2020_92650b2e,meng2022locating} are sensitive to experimental design~\citep{zhang2024bestpracticesactivationpatching}, and ablations can fail to change behavior because redundant pathways compensate (the ``hydra effect''~\citep{mcgrath2023hydraeffectemergentselfrepair}), a case of \textbf{overdetermination}: several independently sufficient causal pathways~\citep{schaffer2003overdetermining}. Subspace patching can produce interpretability illusions by activating alternate pathways~\citep{makelov2023subspacelookingforinterpretability}; circuits often fail to generalize~\citep{li2025interpretationpredictbehaviorunseen} and vary with minor experimental choices~\citep{meloux2025mechanisticinterpretabilitystatisticalestimation}; exhaustive studies on toy models find many incompatible explanations, including for random networks~\citep{melouxeverything}; and causal abstraction methods can produce explanations for random networks in general~\citep{sutter2025nonlinearrepresentationdilemmacausal}.
 
\medskip

\noindent\textbf{Natural language explanations.}
LLM-generated and chain-of-thought explanations are often unfaithful to the computation~\citep{ajwani2024llmgeneratedblackboxexplanationsadversarially,turpin2023languagemodelsdontsay,lanham2023measuringfaithfulnesschainofthoughtreasoning}: many plausible stories can rationalize the same behavior post hoc.

\section{The Deeper Statistical Issue}
\label{sec:issue}
The problems documented in Section~\ref{sec:description} point to a broad statistical fragility. Here, we identify the common structure underlying these failures: the non-identifiability of interpretability queries.
The problems documented in Section~\ref{sec:description} point to a broad statistical fragility. They are, however, of different kinds, and it helps to separate them before identifying the one we consider most fundamental. We distinguish four failure modes (made precise in Section~\ref{sec:stats_inf_view}):
(F1) \textbf{estimation error}: variance across seeds, datasets, and hyperparameters, and missing corrections such as for multiple comparisons;
(F2) \textbf{misspecification}: even the best admissible explanation fits the system poorly;
(F3) \textbf{non-identifiability}: several incompatible explanations fit the system equally well;
(F4) \textbf{estimand/claim mismatch}: the quantity measured is not the one the claim is about, for instance linear decodability of a concept reported as evidence that the network computes it.
Figure~\ref{fig:illus} combines (F1) and (F4). Standard statistical practice addresses (F1), and explicit computational nulls address much of (F4). (F3) is the deeper issue, and we focus on it below.

Behavior-based approaches that study input--output relationships (e.g., feature attributions, behavioral testing) are fundamentally limited by \textbf{underspecification}: multiple, distinct explanations can equally well account for the same input--output patterns \citep{jacovi2021contrastive,rogers2021primer,hagendorff2023machine}. Similar observations in cognitive science motivated the development of brain imaging as a complement to purely behavioral data, with the goal of measuring neural computation and thereby obtaining objective, measurable, and more generalizable quantities \citep{kosslyn1999if,logothetis2008we,churchland1988perspectives}. AI interpretability has followed a related trajectory moving toward analyzing internal computation \citep{mueller2024questrightmediatorhistory}. However, predictive approaches based on internal states (probing, SAEs) inherit standard machine-learning pathologies such as overfitting and poor generalization \citep{belinkov-2022-probing}. These failure modes are instances of \textbf{underspecification}: many predictive models can fit the training data equally well, leaving it unclear which ones posit generalizable causal mechanisms \citep{exp_unders,10.5555/3586589.3586815}.
The manipulability of attributions~\citep{NEURIPS2019_bb836c01} fits the same pattern: inputs are constructed on which the explained output is unchanged while the attribution map changes arbitrarily. If attributions were determined by the queries they are meant to answer, a perturbation preserving those answers would preserve the attribution; manipulability exhibits two explanations of equal fit and very different content.

Causal approaches, introduced in response to the shortcomings of predictive methods, appear at first to provide the scientific rigor needed for generalizable explanations. However, AI systems are large, distributed systems with many interacting components, which gives rise to redundant and context-dependent causal pathways~\citep{FrankleC19}. This creates \textbf{overdetermination}, where multiple distinct causal mechanisms are each independently sufficient to produce the same behavior \citep{doi:10.1073/pnas.91.11.5033,loosemore2012complex,sarkar2022explainableairacemodel}. Then, finding mechanistic stories within complex computational systems can become \textit{too easy}: many different, incompatible explanations can be produced for the same phenomenon \citep{lindsay2023testing,melouxeverything}.
The two sources call for different remedies: underspecification is reduced by restricting the class of admissible explanations (regularization, capacity control, minimality), while overdetermination is reduced by asking richer questions of the system, typically more targeted interventions (Section~\ref{sec:discussion} and Appendix~\ref{app:example}).

\medskip
\noindent\textbf{Identifiability.}
These failure modes can be formalized using the concept of identifiability. Informally, identifiability is the property of a statistical inference task stating that the parameters (explanatory variables) of a statistical model can be uniquely recovered from available observations~\citep{casella2024statistical}. Identifiability is typically a prerequisite for reliable inference in the natural sciences; without it, inferred explanations remain ambiguous. Therefore, substantial work in statistics, unsupervised learning, and causal inference has focused on characterizing identifiability conditions and designing identifiable tasks~\citep{casella2024statistical,Allman2009,Locatello2019,Khemakhem2020,Shpitser2008}.
For interpretability, both underspecification and overdetermination produce non-identifiability, which accounts for several of the observed fragilities:
(i) \textbf{Poor generalization:} when multiple explanations fit the observed data equally well, their explanatory claims can diverge arbitrarily on unseen data. Selecting among these explanations, therefore, depends on arbitrary inductive biases that are rarely validated.
(ii) \textbf{Sensitivity to design choices:} non-identifiability implies a manifold of explanations that achieve a \textit{good fit}. Different algorithmic choices (datasets, optimization procedures, hyperparameters) traverse this manifold differently, and thus produce different explanations.
(iii) \textbf{False discovery:} when explanations are non-identifiable, the probability of recovering a spurious explanation that happens to fit the data increases with the size and complexity of the hypothesis space.

Currently, identifiability is just a conceptual analogy, because interpretability has not yet been formalized as an explicit statistical inference task. Making this formal connection and casting interpretability queries as well-specified statistical estimation problems is a necessary first step toward developing methods whose limitations and assumptions can be explicitly characterized.

\section{The Statistical--Causal Inference Perspective}
\label{sec:stats_inf_view}
A straightforward way to address dead-salmon artifacts across interpretability methods is to compare findings on a trained target network against a randomized alternative: the same architecture with randomized weights analyzed by the same method. This leads to a principled hypothesis test against a null hypothesis of randomized computation, an idea foreshadowed in early work on probing~\citep{conneau-etal-2018-cram,hewitt-manning-2019-structural,ravichander-etal-2021-probing} and in circuit discovery, where \citet{shi2024hypothesistestingcircuithypothesis} give the closest existing operationalization of the testing practice we advocate. We formalize such a test in Appendix~\ref{app:exp} and show that, for probing, it eliminates some false discoveries and substantially reduces effect sizes in standard analyses.
Full weight randomization is only one member of a broader class of \emph{alternative computational hypotheses} that are specific to computational systems and cheap to construct: architecture-matched randomization, block-local reinitialization, random orthogonal transformations of a representation, shuffled labels, task-preserving controls, and partially trained checkpoints. Each answers a different question (``is this structure computed, or already present in the embeddings?'', ``is it specific to this block?''), and interpretability is unusual among empirical sciences in that such nulls can be built directly. They are rarely used in practice.
\begin{wrapfigure}{r}{0.45\linewidth}
    \centering
    \vspace{-6em}
    \includegraphics[width=\linewidth]{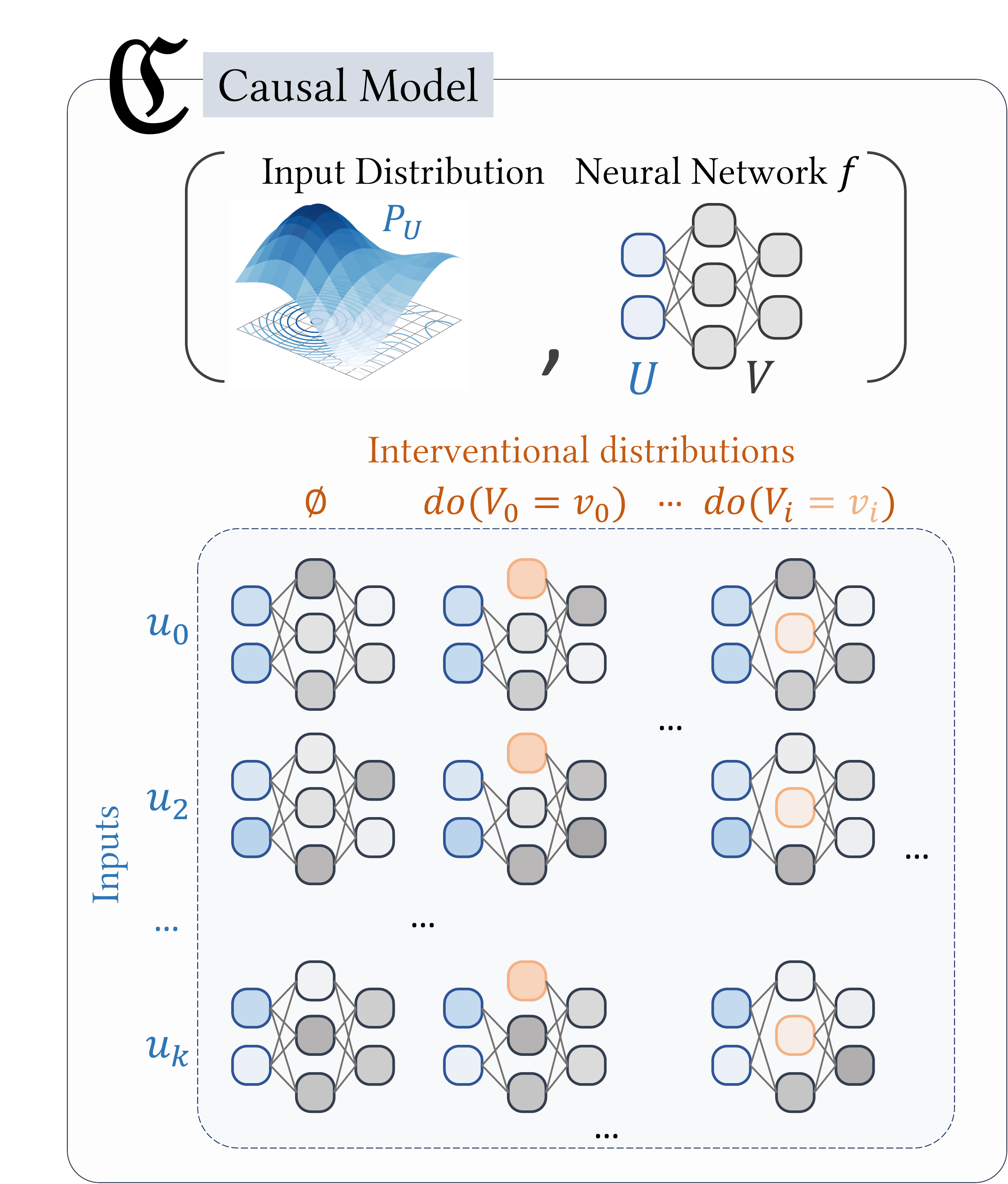}
    \caption{The tuple of a target behavior $P_{\bU}$ and the computational system with its internal components form an SCM.}
    \label{fig:scm}
\end{wrapfigure}

Directly correcting dead-salmon artifacts is, however, a low bar. Hypothesis testing against computational nulls motivates a broader statistical--causal reframing of interpretability, which also lets us address the deeper issue of identifiability.
Here, we sketch one such formalization, viewing interpretability as a problem of \emph{statistical--causal inference}. In this view, explanations are \emph{surrogate models} constructed to answer distributions of causal queries about a computational system. An explanation is useful insofar as it supports prediction and manipulation, generalizes under intervention, and remains robust to noise. This perspective aligns with a growing pragmatist approach to interpretability~\citep{P_ez_2019}. Background on statistical models, estimators, and structural causal models is given in Appendix~\ref{app:background}.

\subsection{Explanations as Surrogate Models}

\paragraph{The system as a causal model.}
Let $f$ be a computational system, typically a neural network, with internal computational elements $\bV$, and let $P_{\bU}$ be an input distribution representing the \emph{behavior of interest} (e.g., arithmetic prompts, images from a domain, factual questions on a topic). As is standard in mechanistic interpretability~\citep{olah2018the,pmlr-v162-geiger22a,geiger2025causalabstractiontheoreticalfoundation}, the pair $\mathfrak{C} := (f, P_{\bU})$ defines a structural causal model (SCM) whose variables are the network's activations and whose causal graph is its computation graph (Figure~\ref{fig:scm}; Appendix~\ref{app:background}). A \emph{causal query} $q(\mathfrak{C})$ is any well-defined quantity about $\mathfrak{C}$, observational or interventional, such as ``What distribution would the network produce if we forced activation $V_i$ to value $v$?''

\begin{wrapfigure}{r}{0.45\linewidth}
    \centering
    \vspace{-1em}
    \includegraphics[width=\linewidth]{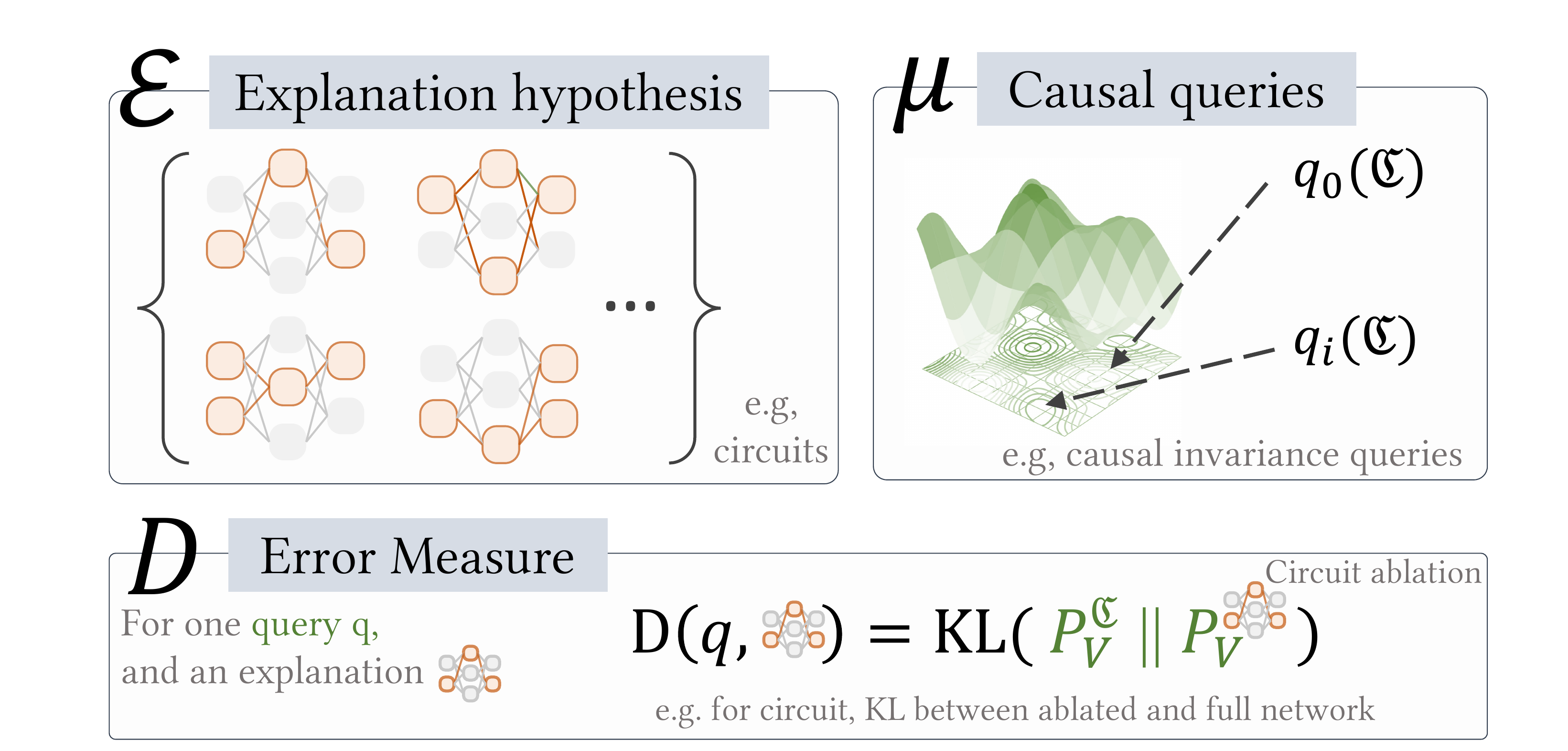}
    \caption{An interpretability task is defined by: $\mathcal{E}$, the hypothesis space; $\mu$, the distribution over causal queries about the SCM; and $D$, the error measure.}
    \label{fig:framework}
\end{wrapfigure}

\paragraph{Surrogates and tasks.}
In an attempt to provide a general statistical-causal perspective on interpretability, we formalize explanations as \emph{surrogate models}: simpler computational descriptions designed to answer chosen collections of causal queries about a target system. This perspective treats interpretability as a form of model compression, where we seek a simpler model that faithfully approximates a complex system's behavior for queries we care about. In this perspective, every interpretability method is characterized by three ingredients:
\begin{enumerate}
  \item \textbf{Query space} $Q$ with \textbf{distribution} $\mu$: The set of causal queries to be answered by the explanation. It dictates what aspects of $\mathfrak{C}$ should be explained. This encodes our explanatory goals.
  \item \textbf{Surrogate set} $\mathcal{E}$: The class of admissible explanations. It dictates what forms the explanation can take, e.g., circuits, sparse subgraphs, linear probes, concept vectors, causal graphs, \dots
  \item \textbf{Discrepancy measure} $D$: How we measure whether a surrogate (member of $\mathcal{E}$) \textit{correctly} answers queries.
\end{enumerate}
This framework is (non-rigorously) illustrated with the example of circuit discovery in Figure~\ref{fig:framework}.
While standard causal inference often concentrates $\mu$ on a \emph{single} causal query (e.g., average treatment effect), interpretability aims to answer \emph{many diverse queries} drawn from a non-trivial distribution $\mu$.
For example, $\mu$ might distribute probability over interventional queries and counterfactual queries across network components, or any functionals of the interventional and counterfactual distributions. 

Each candidate explanation $e \in \mathcal{E}$ defines a \emph{query-answering map} $S_e: Q \to \mathcal{R}$, where $S_e(q)$ is the surrogate's predicted answer to query $q$, and $\mathcal{R}$ is the space of possible answers (e.g., probability distributions, scalar effects, or discrete predictions). The surrogate's \emph{fidelity} is measured by the discrepancy function $D: \mathcal{R} \times \mathcal{R} \to \mathbb{R}_+$ quantifying the error between the answer $q(\mathfrak{C})$ and the surrogate's prediction $S_e(q)$.
Then, we can define the \emph{population risk} of a candidate explanation as the expected error over queries:
\begin{equation}
\label{eq:population_risk}
L_\mu(e) = \mathbb{E}_{q\sim\mu}\big[D\big(q(\mathfrak{C}),\, S_e(q)\big)\big].
\end{equation}
An ideal explanation $e^* \in \mathcal{E}$ minimizes this risk: $e^* \in \arg\min_{e \in \mathcal{E}} L_\mu(e)$. The \emph{residual risk} $\min_{e\in\mathcal{E}} L_\mu(e)$ measures misspecification (F2): if it remains high, no admissible explanation answers the queries well. Debates such as the one on the linear representation hypothesis can be framed in these terms, by estimating the residual risk of the class of linear features.



\paragraph{Interpretability Task and Identifiability.}
We call the triple $(\mu, \mathcal{E}, D)$ an \emph{interpretability task}, fully specifying what we aim to explain (query distribution $\mu$), what explanations are admissible (hypothesis class $\mathcal{E}$), and how we measure success (discrepancy $D$).
The task is \emph{identifiable} if $L_\mu$ admits a unique minimizer in $\mathcal{E}$ (potentially up to predefined acceptable symmetries). Identifiability captures whether the surrogate set can, in principle, be distinguished using the queries deemed relevant by $\mu$. It is a property of the task and the system, not of the network alone, and better uncertainty estimates do not by themselves resolve it. Without identifiability, multiple incompatible explanations achieve the same population risk, and the appropriate output of a method is this equivalence class (in practice, a confidence set over $\mathcal{E}$) rather than one arbitrary member of it. The analysis of Section~\ref{sec:description} indicates that the most common tasks are not identifiable; Appendix~\ref{app:example} works through a minimal example.
Finally, this reframing highlights that explanations are \emph{pragmatic computational summaries} of the structure encoded by $\mathfrak{C} := (f, P_{\bU})$. They are inferred models useful for specific explanatory purposes. 

%

\paragraph{Estimation with Finite Data.}
In practice, we face two types of finite sampling difficulties. First, we observe only finitely many queries $q_1,\ldots,q_n \sim \mu$ from the query distribution. Second, for each query $q_j$, we can only collect a finite amount of computational traces by sampling inputs $\bU \sim P_{\bU}$ and recording the corresponding activations and outputs, potentially under interventions.
Let $\mathcal{T}_n$ denote the complete dataset of query-trace pairs. An interpretability method $M$ acts as an \emph{estimator}, mapping this finite dataset to an estimated surrogate explanation $\hat{e} := M(\mathcal{T}_n).$
A natural estimator is given by empirical risk minimization: choose $\hat{e}$ to minimize the empirical risk $\widehat{L_\mu}(e) = \frac{1}{n}\sum_{j=1}^n \widehat{D}(q_j(\mathfrak{C}) ,\, S_e(q_j))$, where $\widehat{D}$ is estimated from finite traces.
%

\paragraph{Existing criteria as special cases.}
Several established criteria are instances of $L_\mu$ (Table~\ref{tab:interpretability_methods}). Probe accuracy corresponds to an observational $\mu$, linear maps for $\mathcal{E}$, and a classification loss. Circuit faithfulness corresponds to $\mathcal{E}$ the subgraphs, $\mu$ the ablation of each candidate's complement, and $D$ the KL divergence between outputs. \citet{shi2024hypothesistestingcircuithypothesis} add hypothesis tests on top of such criteria for circuits; our framework extends that testing practice to other choices of $(\mu,\mathcal{E},D)$. Interchange-intervention accuracy (IIA) in causal abstraction~\citep{geiger2025causalabstractiontheoreticalfoundation} is exactly $1-L_\mu$ for $\mu$ uniform over interchange interventions, $\mathcal{E}$ the alignments between a high-level causal model and neural subspaces, and $D$ the 0--1 loss on counterfactual predictions.
Relative to these criteria, the framework adds three things. First, $\mu$ is declared rather than fixed by convention; since different choices of $\mu$ induce different equivalence classes, the choice affects the answer. Second, it asks for uncertainty and nulls: IIA, for instance, is estimated from a finite sample of interchange pairs and often a stochastic alignment search~\citep{pmlr-v236-geiger24a}, yet is typically reported as a point estimate. Third, it raises the question of unique recoverability: causal abstraction characterizes when an abstraction relation holds, but not when the abstraction can be uniquely recovered from a finite family of interventions.
Other prior work also fits in the framework: \citet{pmlr-v202-senetaire23a} proposed a statistical framing for feature attribution, \citet{bordt2025position} argue for descriptive statistics of functionals, and \citet{meloux2025mechanisticinterpretabilitystatisticalestimation} study the variance of circuit discovery.

\section{Practical Recommendations}
\label{sec:rec}
To move from anecdotal discovery to cumulative science, interpretability research should make explicit what is being explained, under which assumptions, and with what uncertainty.
%
Drawing on the philosophy of science~\citep{chang2004inventing,Woodward2003-WOOMTH,potochnik2017idealization} and recent calls for a pragmatic approach to interpretability~\citep{davies2024cognitiverevolutioninterpretabilityexplaining,williams2025mechanisticinterpretabilityneedsphilosophy}, the framework proposed above naturally distinguishes the \emph{explanandum} (what is to be explained, encoded in $\mu$) from the \emph{explanans} (what does the explaining, encoded in $\mathcal{E}$). Researchers and practitioners have substantial freedom in choosing both. There is no single ``correct'' explanation of a neural network. The appropriate type of description depends on one's purposes~\citep{potochnik2017idealization}. Descriptive understanding corresponds to queries about observational distributions; predictive goals involve queries requiring surrogates to generalize to new input distributions; control and intervention require queries about counterfactual or interventional distributions.

However, once the explanatory project is specified, i.e., once $(\mu, \mathcal{E}, D)$ are fixed, the explanation becomes an \textbf{objective inference problem}. The best surrogate $e^* \in \mathcal{E}$ is the one minimizing $L_\mu(e)$, and is a property of the system itself and the interpretability task. If the task is identifiable, this explanation is unique (up to permissible symmetries, e.g., rotation invariance in representation space). This reconciles pluralism about explanatory goals with rigor about explanatory claims.

Perhaps most critically, the statistical framing demands that interpretability methods report not just point estimates but \emph{confidence sets} or \emph{posterior distributions} over explanations (in case of Bayesian framing). Just as we would not trust a clinical trial reporting effect sizes without confidence intervals, we may not trust interpretability claims without uncertainty quantification. 

\paragraph{Concrete workflow.}
Authors should \textbf{formalize} the explanatory target by specifying the behavior and input distribution being explained, the query space $\mu$, the explanation class $\mathcal{E}$, and the discrepancy criterion $D$. They should report the \textbf{residual risk} of the selected explanation, which indicates whether $\mathcal{E}$ is adequate at all. They should \textbf{analyze identifiability} by determining whether the available evidence supports a unique explanation or only an equivalence class of circuits, features, abstractions, or mechanisms, and in the latter case report the class rather than one member. They should \textbf{estimate with uncertainty} by reporting confidence sets, posterior uncertainty, bootstrap variability, or sensitivity analyses over seeds, datasets, checkpoints, interventions, and analysis choices. They should \textbf{validate} the claim against explicit computational nulls (Section~\ref{sec:stats_inf_view}), such as architecture-matched randomization, block-local reinitialization, random orthogonal transformations, shuffled labels, task-preserving controls, or partially trained checkpoints. Finally, they should \textbf{report} the scope of the claim, including whether it is descriptive, predictive, causal, or mechanistic, which assumptions it relies on, and which alternatives remain unresolved.
The central question should not be whether an experimental finding looks explanatory, but what estimand it targets, under what assumptions it is identifiable, what is the uncertainty, and which alternative computational hypotheses it rules out (and which ones it does not).

We mean for the reporting commitments to be taken literally: nulls, uncertainty, and scope should be stated. The formalism, however, is meant as a design tool rather than as notation every paper must use. A paper need not write out $(\mu,\mathcal{E},D)$ formally, but its authors should be able to say what they are if asked; Table~\ref{tab:interpretability_methods} shows that existing methods already make these choices implicitly.

\paragraph{Choosing the task.}
No triple $(\mu,\mathcal{E},D)$ is correct independently of the research purpose, and the framework does not prevent choosing one that makes a favored explanation come out on top. What it requires is that the choice be reported, so that readers can see, for example, that $\mathcal{E}$ was restricted to a short list of candidates and treat this as an assumption rather than a finding. Such restrictions already occur in practice, implicitly, through the choice of method. Identifiability cannot be obtained by narrowing $\mu$: in the realizable case, removing queries from $\mu$ can only enlarge the set of exact minimizers (Appendix~\ref{app:example}).

\paragraph{Cost.}
These practices need not be expensive, and apply to large models. For post-hoc methods no retraining is needed, and several nulls cost a single forward pass or less (random orthogonal rotations of the residual stream at the studied layer, block-local reinitialization, shuffled-label or permuted-input controls). Bootstrapping over queries and traces reuses cached activations. The number of null draws $B$ need not be large: the Monte Carlo test of Appendix~\ref{app:exp} is valid for any $B$, which only sets its resolution (the smallest attainable $p$-value is $1/(B+1)$, already below $0.05$ for $B=20$).

\section{Alternative Views}
\label{sec:alternatives}
We consider five positions that differ from ours in some respect.

\paragraph{Better methods, not statistics.}
One may hold that the failures of Section~\ref{sec:description} are method-specific problems that better methods will fix. Many of the fixes proposed so far (capacity-controlled probes, amnesic probing, cross-validated SAEs, random baselines) are, however, statistical remedies introduced case by case. Naming them as such makes it easier to apply them consistently and to see what they leave unaddressed.

\paragraph{Mechanistic realism.}
On a realist view, a network implements a particular algorithm, and causal abstraction defines what it means for a high-level model to be implemented~\citep{geiger2025causalabstractiontheoreticalfoundation}. Our argument does not require rejecting this view. A realist who wants to recover the true algorithm needs the evidence to single it out, which is precisely the identifiability question; identifiability is the empirical precondition for realist claims.

\paragraph{Explanatory virtues and simplicity priors.}
When several hypotheses fit the data, scientists rely on simplicity and other explanatory virtues to choose between them, and \citet{ayonrinde2025mathematical,ayonrinde2025evaluating} develop such criteria for mechanistic interpretability. We partly adopt this view: in our notation, virtues enter as restrictions of $\mathcal{E}$ or as priors over it. They do not always single out an explanation, however. Many incompatible circuits remain after selecting for both faithfulness and minimality~\citep{melouxeverything}, and in the example of Appendix~\ref{app:example} the two minimal explanations are exchanged by a symmetry of the model, so no symmetric preference can choose between them.

\paragraph{Applied statistics without causality.}
\citet{bordt2025position} argue for a statistical view of explanations as descriptive statistics of a function. For observational queries this is compatible with our framework. Many interpretability claims are, however, interventional (``this component implements this step''), and evaluating them requires the causal layer.

\paragraph{Pragmatic engineering.}
A pragmatic view judges explanations, or inherently interpretable models~\citep{rudin2019stop}, by their usefulness on concrete tasks. We share this pragmatism. A utility benchmark without a null or an uncertainty estimate can still return dead salmons, though, so the statistical commitments apply to it as well.

\paragraph{Where the framework does not apply.}
The framework is not meant for exploratory work, whose purpose is to generate hypotheses rather than to test them; for studies of human-facing usability, where the target is what people understand rather than fidelity to the system; or for systems that cannot be modeled as an SCM with a fixed input distribution $P_{\bU}$.

\section{Discussion}
\label{sec:discussion}
The systematic failures documented in Section~\ref{sec:description} demanded an explanation. We have argued that these pathologies share a common root cause: \textbf{non-identifiability}. Most current interpretability tasks attempt to infer explanatory structures that are not uniquely determined by available computational traces.
To trace a path forward, we proposed one formalization of interpretability as statistical-causal inference. This framework is tentative rather than definitive; we encourage the community to improve upon it. The important aspect is the \emph{methodological commitment} to making assumptions explicit and quantifying uncertainty rigorously. The framework permits interpreting previous failures (see Appendix~\ref{app:interp} for more details). While it does not remove non-identifiability where the task is non-identifiable, it makes the equivalence class visible instead of letting a method report one arbitrary member of it, and makes explicit which changes to the task would shrink it.

\subsection{Towards Useful and Identifiable Interpretability Tasks}

Identifiability is not an intrinsic property of the model under study but of the interaction between $\mu$, $\mathcal{E}, D$, the model $f_{NN}$, and the behavior of interest $P_{\bU}$. We might wonder what choices to make in order to improve the identifiability and usefulness of interpretability queries.

\textbf{Query richness.} The queries in the support of $\mu$ must be sufficiently discriminative to distinguish candidate explanations in $\mathcal{E}$. There is a fundamental trade-off between discriminative power and sample efficiency. If $\mu$ spreads probability mass over a large support, accurately estimating $L_\mu$ may require prohibitive amounts of interventional data. Conversely, concentrating $\mu$ on too few queries risks not singling out one explanation in $\mathcal{E}$. Enriching $\mu$ is the main remedy for overdetermination. Interchange interventions are one such enrichment, which is why interchange-intervention accuracy discriminates better than ablation-based faithfulness (Appendix~\ref{app:example}). IIA still scores a given hypothesis rather than ranking the hypothesis space: high IIA for one explanation does not preclude high IIA for an incompatible one~\citep{melouxeverything,sutter2025nonlinearrepresentationdilemmacausal}, and with expressive alignment maps high IIA can be reached without the network implementing the hypothesis~\citep{makelov2023subspacelookingforinterpretability}. It also depends on intervention design~\citep{zhang2024bestpracticesactivationpatching}; when redundant pathways are context-dependent, we do not know in advance which intervention family will separate them.

\textbf{Expressivity vs. parsimony in $\mathcal{E}$}. Conversely, the hypothesis class must have sufficient capacity to approximate the queries well (low bias) but not so much flexibility that many distinct explanations all achieve low error (large equivalence classes, high variance, non-identifiability). This is akin to the classical bias-variance tradeoff, pointing toward standard fixes like regularization of the hypothesis class~\citep{belinkov-2022-probing}. Restricting $\mathcal{E}$ is the main remedy for underspecification. Simplicity priors and other explanatory virtues~\citep{ayonrinde2025evaluating} are legitimate ways of doing so, but, as discussed in Section~\ref{sec:alternatives}, they do not always single out one explanation.

\textbf{Human cognitive constraints.} Interpretability is meant to facilitate \emph{human understanding}. Empirical studies suggest people can mentally simulate models with only a handful of interacting components~\citep{lombrozo2006structure,wilkenfeld2013understanding,keil2006explanation,hassija2024interpreting}. Explanations exceeding these structural limits may be technically \textit{correct} yet fail to provide insight. Designing $\mathcal{E}$ with human simulability in mind ensures that understanding remains the end goal.

\subsection{Opportunities for Future Work}

\paragraph{Characterizing identifiability conditions.}
A systematic theoretical program could characterize when specific $(\mu, \mathcal{E}, D)$ triplets are identifiable, mirroring similar efforts in causal inference~\citep{Shpitser2008} and unsupervised learning~\citep{Locatello2019,Khemakhem2020}. What symmetries and invariances are unavoidable in representation space, and when is identifiability up to such equivalences acceptable? Constructing a taxonomy of identifiable interpretability tasks would provide actionable guidance for practical scenarios. One route is the following. In the realizable case, where some $e^*\in\mathcal{E}$ answers all queries exactly, identifiability reduces to whether a family of evaluation functionals separates a function class (under misspecification, to uniqueness of a projection). Since identifiability is monotone in $\mu$ and anti-monotone in $\mathcal{E}$, tasks can be ordered along these two axes. For a given $\mathcal{E}$, a taxonomy could report the equivalence class induced by a standard $\mu$, the minimal enrichment of $\mu$ that collapses it, and the sample cost of that enrichment. Interpretability is in a favorable position for this program compared with other sciences: the system is fully accessible, arbitrary interventions are possible, and nulls can be constructed. This also raises identifiability questions that do not arise elsewhere.

\paragraph{Bayesian interpretability and uncertainty quantification.}
Bayesian approaches offer an elegant framework for handling non-identifiability and quantifying uncertainty~\citep{gelman2013bayesian}. Specifically, one could specify a \textbf{prior distribution} $\pi(e)$ over the explanation class $\mathcal{E}$, encoding structural preferences (e.g., sparsity, modularity) or incorporating prior information from related studies. Then, the \textbf{likelihood model} $P(\mathcal{T}_n \mid e)$ describes how computational traces are generated given explanation $e$. Finally, the \textbf{posterior updates} via Bayes' rule: $\pi(e \mid \mathcal{T}_n) \propto P(\mathcal{T}_n \mid e) \pi(e)$, refines beliefs as observations accumulate. Then, \textbf{credible sets} can quantify uncertainty.
When explanations are non-identifiable, the posterior remains diffuse across an equivalence class; uncertainty quantification naturally reflects this fundamental ambiguity. Conversely, as more discriminative queries are observed, the posterior concentrates. This further provides a principled framework for active setup: strategically selecting queries from $\mu$ that maximally reduce posterior uncertainty.

\paragraph{Meta-analysis and cumulative science.}
Meta-analytic methods~\citep{borenstein2021introduction} could coherently aggregate evidence across studies, accounting for heterogeneity in $\mu$, $\mathcal{E}$, and experimental conditions. Standardized effect size measures, pre-registration of analyses, and open sharing of collected computational traces would enable interpretability to become a cumulative science where knowledge systematically builds over time. In general, responses proposed by other fields~\citep{poldrack2017scanning,korbmacher2023replication} become available for interpretability.

\section*{Acknowledgements}

This work was partially conducted within French research unit UMR 5217 and was partially supported by CNRS (grant ANR-22-CPJ2-0036-01) and by MIAI@Grenoble-Alpes (grant ANR-23-IACL-0006). It was granted access to the HPC resources of IDRIS under the allocation 2025-AD011014834 made by GENCI.\\

\clearpage
\bibliographystyle{plainnat}
\bibliography{causal_abs,anthology}


\clearpage
\appendix

\section{Fixing Dead Salmons with Hypothesis Testing}
\label{app:exp}

Consider an interpretability method $M$ that aims to explain a neural network $f$ for some input behavior $P_{\bU}$, we note $\mathfrak{C}$ the tuple $(f, P_{\bU})$ as done in the main paper. The method produces an explanation $\hat{e}$ from finite observations from $\mathfrak{C}$, possibly under interventions. This tentative explanation could take the form of a circuit, a set of important features, concept activation vectors, or any other hypothesis class. We might wonder how to prevent dead salmon artifacts from arising with the interpretability method $M$? 

A simple, direct, and natural solution is to frame this question as a hypothesis test against a null hypothesis where the observed explanation arises from random computation. Already, for probing, \citet{ravichander-etal-2021-probing} discusses the possibility of comparing the probe against a probe trained on random embeddings. The general idea is to construct a family of null models represented by a distribution $P_{\tilde{\mathfrak{C}}}$, which preserves the network’s architectural properties while disrupting the specific computational mechanisms we aim to explain. Such null models can be obtained, for example, via full weight randomization, block-local reinitialization, random orthogonal transformations of representations, partially trained checkpoints, or label shuffling (recovering the standard permutation test). Different nulls answer different questions, and the choice should be reported.

For a given interpretability method, we define a test statistic $T(\hat{e}, \mathfrak{C})$ that quantifies explanatory fit for the interpretability task at hand. For example, for probing methods, $T$ could be test accuracy; for circuit discovery, $T$ could measure behavioral fidelity; for attribution methods, $T$ could quantify the correlation between attribution scores and actual intervention effects.

Applying the intepretability method $M$ to one null model $\tilde{\mathfrak{C}}^{(b)}$ from the randomized family yields explanations $\tilde{e}^{(b)} = M(\tilde{\mathfrak{C}}^{(b)})$ and corresponding null statistics $T_{\text{null}}^{(b)} = T(\tilde{e}^{(b)}, \tilde{\mathfrak{C}}^{(b)})$. Then, following standard procedure, the Monte Carlo estimated $p$-value is:
\begin{equation}
\label{eq:mc_pvalue}
\hat{p} = \frac{1 + \sum_{b=1}^B \mathbb{I}\{T_{\text{null}}^{(b)} \ge T_{\text{obs}}\}}{B + 1},
\end{equation}
where $T_{\text{obs}} = T(\hat{e}, \mathfrak{C})$. The addition of 1 to both the numerator and denominator ensures Type~I error control:
$
\Pr(\hat{p} \le \alpha \mid H_0) \le \alpha,
$ where $H_0$ is the null hypothesis~\citep{north2002note,phipson2010permutation}. This guarantee holds for any number of null draws $B$; $B$ only sets the power and resolution of the test, the smallest attainable $p$-value being $1/(B+1)$ (about $0.048$ for $B=20$ and $0.0099$ for $B=100$). By design, when the randomization includes full weight reinitialization, no dead salmon artifacts can remain.

\subsection{Experiments}
To illustrate the hypothesis test, we experiment with three probing tasks.

\begin{figure*}
    \centering
    \includegraphics[width=0.8\linewidth]{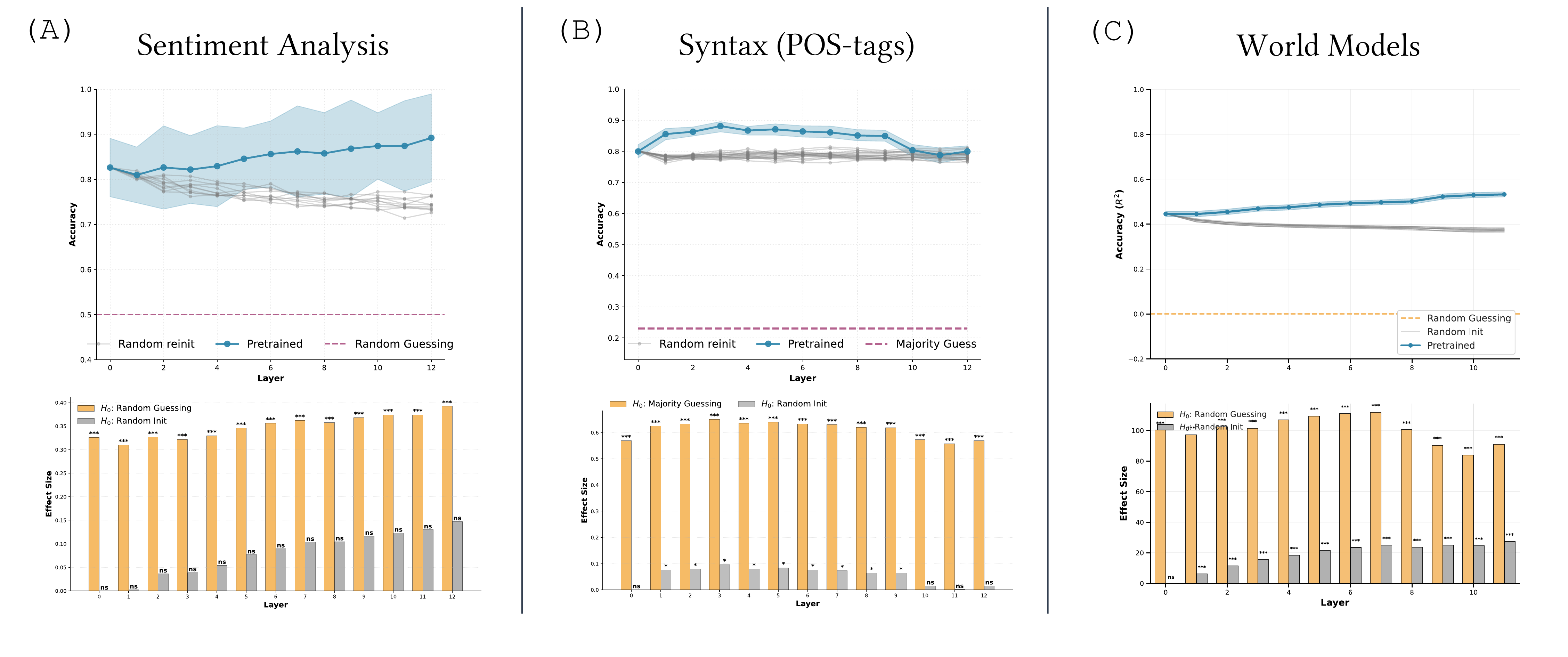}
    \caption{(A) Sentiment analysis experiment where probes on pretrained BERT are compared against probes trained on randomly reinitialized models. (B) Same experiment based on predicting syntactic labels (POS tags). (C) Reproducing the first experiment of Table 2 in \cite{gurnee2024language}, probing for indications of world models on pythia-160m.}
    \label{fig:experiments}
\end{figure*}

\textbf{Sentiment Analysis (IMDb).}
We reuse the IMDB sentiment classification setup from Figure~\ref{fig:illus}. For each layer of \texttt{BERT-base-uncased}, we extract the average sentence embedding and train a linear probe to predict binary sentiment. We also train probes on $B{=}100$ random reinitializations of the model, and evaluate statistical significance using the hypothesis test described above. All probes are trained and evaluated on 1000 sentences with 10-fold cross-validation.
Figure~\ref{fig:experiments}(A) reports (i) the average probe accuracy at each layer for the pretrained model, the randomized models, and a random guessing baseline, and (ii) the corresponding effect sizes relative to random guessing and to randomized models. While all pretrained layers outperform random guessing with large effect sizes, none are statistically distinguishable from the random reinitializations under the new test. Later layers, however, show a clear upward trend in effect size relative to randomized models.

\textbf{Syntactic Structure (POS Tagging).}
We next assess token-level syntactic information using POS-tagging probes~\citep{tenney-etal-2019-bert}. For each layer of \texttt{BERT-base-uncased}, we extract contextual token embeddings and train logistic regression probes on a subset of CoNLL-2003, one probe per layer that should work for all tokens and all POS tags. As above, we also train probes on $B{=}100$ random reinitializations and apply the same statistical test. Probes are evaluated with 10-fold cross-validation on 500 sentences.
Figure~\ref{fig:experiments}(B) reports the layer-wise probe accuracy and effect sizes relative to a majority baseline and to randomized models. Consistent with prior work, POS accuracy peaks in middle layers~\citep{tenney-etal-2019-bert}. However, when tested against randomized models rather than random guessing, only the middle layers remain statistically above the null, and the effect sizes are substantially reduced. This should not be read as saying that the other layers lack POS information: much of it is already linearly available from the token embeddings and survives a pass through random layers. A random-guessing baseline conflates structure computed by the model with structure already present in the input; the architecture-matched null separates the two, and the quantity of interest is the pretrained model's increment over it. A task whose target is not linearly decodable from the tokens would make a useful contrast.

\textbf{World Models (Space and Time).}
Finally, we investigate the emergence of linear representations of space using the ``world places'' dataset from~\citep{gurnee2024language}. Using \texttt{pythia-160m}, we extract average residual stream activations on each token of place names and train linear ridge regression probes to predict their geospatial coordinates (latitude and longitude). We compare the pretrained model against $B=100$ baselines where transformer block weights are randomized while embeddings remain fixed. Probes are evaluated using $R^2$ scores with 10-fold cross-validation.
Figure~\ref{fig:experiments}(C) reveals that raw embeddings (Layer 0) contain latent spatial structure ($R^2 \approx 0.45$), significantly outperforming random guessing ($Z \approx 100$). Passing these embeddings through randomized transformer blocks decreases linear readout (to $R^2 \approx 0.38$ at the last layer). In contrast, the pretrained model's layers slightly improve this spatial linearity relative to the random baseline, suggesting that deeper layers progressively construct a more coherent spatial representation. By the final layers, the learned structure statistically surpasses the random baseline ($Z \approx 25$), confirming that the model eventually learns to encode space explicitly beyond the geometry inherent in the embeddings.


\begin{table*}[ht]
\centering
\renewcommand{\arraystretch}{1.25}
\begin{tabular}{@{}p{3.2cm}p{5.1cm}p{5.1cm}@{}}
\toprule
\textbf{Method} & \textbf{Hypothesis space $\mathcal{E}$ (surrogate model)} & \textbf{Typical causal query $q(\mathfrak{C})$ and error criterion $D$} \\
\midrule
\textbf{Performance benchmarking} 
& Single scalar summarizing predictive performance (e.g., accuracy, calibration, perplexity).  
& \textit{Observational query:} output score distribution under $P_{\bU}$.  
Error: difference in expected performance metrics, e.g., $|{\mathbb E}[f(\bU)] - {\mathbb E}[\hat{f}(\bU)]|$. \\ 
\midrule
\textbf{Probing (linear / diagnostic classifiers)} 
& Linear or shallow classifiers mapping internal activations to target variables (e.g., part-of-speech tags).  
& \textit{Observational query:} conditional distribution $P(Y\,|\,\text{activations})$.  
Error: classification loss. \\ 
\midrule
\textbf{Feature attributions (saliency, SHAP, Integrated Gradients)} 
& Input-level additive surrogates assigning contribution scores so that $f(x)$ is approximated by $\sum_i e_i(x)\,x_i$ relative to a baseline.  
& \textit{Counterfactual queries:} local (additive) approximation of model behavior around inputs $x$.  
Error: fidelity loss between model predictions and surrogate reconstruction \\ 
\midrule
\textbf{Concept-based methods (e.g., TCAV, ACE, concept bottleneck models)} 
& Surrogates mapping internal activations to interpretable concept variables and modeling $f$’s dependence on them.  
& \textit{Interventional queries:} model sensitivity or dependence on interpretable concept activations within latent space.  
Error: deviation between surrogate-predicted and model-predicted sensitivities (e.g., directional derivative mismatch). \\ 
\midrule
\textbf{Circuit discovery} 
& Subgraphs of the computational graph representing causal mechanisms.  
& \textit{Interventional queries:}  outputs distribution under targeted ablations encoded by the circuit.  
Error: consistency in output distribution, e.g.,
$KL\big(P_{\mathfrak{C}}(Y \mid U) \, \| \, P_{\text{circuit}}(Y \mid U)\big)$; \citet{shi2024hypothesistestingcircuithypothesis} add hypothesis tests on such criteria. \\ 
\midrule
\textbf{Causal tracing (patching, mediation analysis)} 
& Scalar importance scores over units or connections inferred from intervention or mediation effects.  
& \textit{Counterfactual mediation queries:} total, direct, or indirect effect of node $V_i$ on target $Y$.  
Error: causal abstraction error, measuring causal alignment as violation of commutative properties of abstraction and intervention. With $\mu$ uniform over interchange interventions and $D$ the 0--1 loss, $1-L_\mu$ is interchange-intervention accuracy. \\ 
\midrule
\textbf{Causal abstraction / model-level alignment} 
& High-level structural causal model with mappings from low-level network variables $\bV$ to abstract variables $\bZ$.  
& \textit{Interventional invariance queries:} the actions of abstracting from $V$ to $Z$ and intervening should commute.
Error: causal abstraction error, measuring causal alignment as violation of commutative properties of abstraction and intervention. \\ 
\bottomrule
\end{tabular}
\caption{Interpretability methods as instances of the statistical--causal framework of \emph{surrogate models}. Each method specifies a hypothesis class $\mathcal{E}$, causal query family $q(\mathfrak{C})$, and associated error measure $D$ quantifying how faithfully the surrogate answers the queries.}
\label{tab:interpretability_methods}
\end{table*}
\section{A Minimal Worked Example: Two Sufficient Pathways}
\label{app:example}

We give a minimal instance of overdetermination in which standard circuit faithfulness is not identifiable, a minimality constraint does not restore identifiability, and enriching $\mu$ with single-component interventions does.

\paragraph{System.}
Let $X\in\{0,1\}$ with $\Pr(X=1)=p\in(0,1)$. Two internal nodes copy the input, $A := X$ and $B := X$, and the output is $Y := A \lor B$. The model thus computes $Y=X$ through two independently sufficient pathways.

\paragraph{Surrogates.}
Let $\mathcal{E}=\{\emptyset,\{A\},\{B\},\{A,B\}\}$ be the circuits over the internal nodes. The surrogate $S_e$ associated with $e\in\mathcal{E}$ is the model in which the nodes outside $e$ are ablated to $0$. A surrogate answers an intervention on a node in $e$ by setting that node, and an intervention on a node outside $e$ has no effect, since that node is ablated. We take $D$ to be the total variation distance between output distributions, averaged over $X$ (KL divergence leads to the same conclusions).

\paragraph{Standard faithfulness.}
Let $\mu_0$ put all mass on the observational query $q_0$: the output distribution given $X$. Ablating the complement of $\{A\}$ sets $B=0$, so $Y=A=X$; similarly for $\{B\}$; $\{A,B\}$ ablates nothing; and $\emptyset$ gives $Y=0$. Hence
\[
L_{\mu_0}(\{A\}) = L_{\mu_0}(\{B\}) = L_{\mu_0}(\{A,B\}) = 0, \qquad L_{\mu_0}(\emptyset)=p.
\]

\begin{proposition}
The task $(\mu_0,\mathcal{E},D)$ is not identifiable: $\arg\min_{e} L_{\mu_0}(e)=\{\{A\},\{B\},\{A,B\}\}$.
\end{proposition}

\begin{proposition}
Restricting to minimal circuits does not restore identifiability. More generally, no preference over $\mathcal{E}$ that is invariant under exchanging $A$ and $B$ can select a unique explanation.
\end{proposition}
\begin{proof}
The minimal-cardinality minimizers are $\{A\}$ and $\{B\}$. The map $\sigma$ exchanging $A$ and $B$ leaves the model unchanged and maps $\{A\}$ to $\{B\}$. Since $\mu_0$ is $\sigma$-invariant, $L_{\mu_0}(\{A\})=L_{\mu_0}(\{B\})$, and any $\sigma$-invariant prior or penalty assigns them the same weight.
\end{proof}

\paragraph{Enriched queries.}
Let $\mu_1$ put positive mass on $q_0$ and on the interventional queries $q_{A,a}$ and $q_{B,b}$, the output distribution given $X$ under $\mathrm{do}(A=a)$ or $\mathrm{do}(B=b)$, for $a,b\in\{0,1\}$. Under $\mathrm{do}(A=1)$, the model outputs $Y=1$ for all $X$. The surrogate $\{B\}$ ignores interventions on $A$ and predicts $Y=X$, which is wrong when $X=0$; symmetrically for $\{A\}$ under $\mathrm{do}(B=1)$. The surrogate $\{A,B\}$ reproduces the model exactly, and $\emptyset$ already errs on $q_0$.

\begin{proposition}
The task $(\mu_1,\mathcal{E},D)$ is identifiable, with unique minimizer $\{A,B\}$: $L_{\mu_1}(\{A\})$ and $L_{\mu_1}(\{B\})$ are at least $\mu_1(q_{B,1})(1-p)$ and $\mu_1(q_{A,1})(1-p)$ respectively, which are positive, while $L_{\mu_1}(\{A,B\})=0$.
\end{proposition}

\paragraph{Remarks.}
(i) Setting $A$ to $1$ is setting it to the value it takes on the source input $X'=1$, i.e., an interchange intervention; the enrichment above is the one performed by interchange-intervention accuracy, which is why IIA separates the candidates here.
(ii) Under $\mu_0$, the appropriate report is the equivalence class (``either pathway suffices; both are consistent with the data''), not one of its members.
(iii) Monotonicity in $\mu$: in the realizable case, the set of exact minimizers is $Z(\mu)=\{e: S_e(q)=q(\mathfrak{C})\ \text{for $\mu$-almost every } q\}$. If $\operatorname{supp}\mu\subseteq\operatorname{supp}\mu'$, then $Z(\mu')\subseteq Z(\mu)$. Adding queries can only shrink the equivalence class, and removing queries can only enlarge it.
(iv) Here we knew which interventions separate the pathways. In larger networks, where redundant pathways can be context-dependent, finding such a family is itself the open problem (Section~\ref{sec:discussion}).
\section{Re-Interpreting Documented Issues}
\label{app:interp}
The framework does not prescribe which $(\mu, \mathcal{E}, D)$ researchers should adopt. Rather, it provides a \emph{shared language} for making assumptions explicit and rooted in the tools of statistical inference. Different research programs will choose different hypothesis classes or query distributions; the framework ensures that such choices are transparent and their implications are analyzable. Table~\ref{tab:interpretability_methods} in the appendix illustrates how existing interpretability methods can be mapped into this formulation, each implicitly making assumptions about queries, surrogates, and error metrics. The goal is to reinterpret and compare these methods, not to replace them. Under this view, many of the issues described in Section~\ref{sec:description} can be understood as problems of \emph{non-identifiability} (F3), alongside estimation error (F1), misspecification (F2), and estimand/claim mismatch (F4).

\paragraph{Behavioral Benchmarks.}
Benchmarks that evaluate model outputs against a gold standard (averaged success over input distributions) ask an identifiable question: \textit{how well does the model perform under a specific task distribution and error metric?} This is arguably the simplest form of interpretability and drove most of the progress in AI. Its usefulness depends on the construction of the benchmark, but the inference problem is well-posed.

\paragraph{Concept-Based Approaches.}
Predictive methods (probes, SAEs) inherit non-identifiability issues from the underlying underspecification of machine learning tasks~\citep{10.5555/3586589.3586815}.
For example, methods like Concept Activation Vectors~\citep{kim2018interpretability, cunningham2023sparseautoencodershighlyinterpretable} postulate that internal states $\bv$ are generated by interpretable concepts $\mathbf{z}$ via $\bv = g(\mathbf{z})$.
This is an instance of (causal) representation learning, which is \textbf{non-identifiable} without auxiliary information \citep{Locatello2019, Khemakhem2020}. 
It is therefore unsurprising that proposed improvements mirror standard remedies for underspecification in machine learning: regularization in the form of capacity control for probes~\citep{belinkov-2022-probing} or cross-validation to assess generalization~\citep{kantamneni2025sparseautoencodersusefulcase}.

\paragraph{Causal Mediation Analysis.}
Methods like causal mediation analysis \citep{meng2022locating, vig2020causalmediationanalysisinterpreting} estimate the indirect effect of a component on observed outputs. As the intervention and model are fully specified, the mediation estimand is unique and \textbf{identifiable}. However, the explanatory claim that a component with high effect is the \textit{locus} of a mechanism is \textbf{not identifiable}, because of the overdetermined causal structure. 

\paragraph{Circuit Discovery and Causal Approaches.}
Circuit discovery seeks a subgraph $G' \subset G$ that preserves the model's performance. This task faces the "Hydra effect" \citep{mcgrath2023hydraeffectemergentselfrepair} and causal overdetermination. If parallel pathways $A$ and $B$ are sufficient, circuits containing only $A$ or only $B$ both satisfy fidelity criteria. Appendix~\ref{app:example} makes this precise and shows that adding single-component interventions to $\mu$ restores identifiability in that case. Thus, even \textit{correct} causal methods may recover many different explanations consistent with the same behavior \citep{melouxeverything}. Addressing this requires formulating identifiable causal questions. Causal abstraction \citep{Beckers_Halpern_2019,geiger2025causalabstractiontheoreticalfoundation} offers a promising direction, as it operates at a coarser representational level where overdetermination can be absorbed into the abstracted representations. However, current operational metrics demonstrate empirical non-identifiability~\citep{melouxeverything, sutter2025nonlinearrepresentationdilemmacausal}.

\section{Background: Statistical--Causal Inference}
\label{app:background}

Statistical inference provides the rigorous framework through which empirical observations become scientific knowledge~\citep{cox2006principles,lehmann1998theory}. We argue that interpretability, like every empirical science, must be grounded in these principles. This appendix gives the background used in Section~\ref{sec:stats_inf_view}.

\paragraph{Statistical Models and Identifiability.}
A \emph{statistical model} is a family of probability laws $\{\mathbb{P}_\theta^{\bV}:\theta\in\Theta\}$ on a sample space $\mathcal{V}$, indexed by parameters $\theta\in\Theta$. Here, $\bV$ denotes observed data. Intuitively, we assume data arises from some process indexed by unknown parameters~$\theta$, and the goal is to recover~$\theta$ from observations.
Sound inference requires \emph{identifiability}: distinct parameters must induce distinct distributions over observables. Formally, a model is identifiable if $\theta \neq \theta' \implies \mathbb{P}_\theta^{\bV} \neq \mathbb{P}_{\theta'}^{\bV}$. Without identifiability, hypotheses cannot be distinguished from data, rendering inference ill-posed. 

\paragraph{Estimators and Uncertainty Quantification.}
Given finite observations $\mathcal{D}_n=\{\bv^{(i)}\}_{i=1}^n$, an \emph{estimator} $T$ produces an estimate $\hat{\theta} := T(\mathcal{D}_n)$ of unknown parameters $\theta$. Its quality can be assessed through various statistical properties: (i) \textbf{Bias}: Does it recover the correct parameter on average? (ii) \textbf{Variance}: How much does the estimate vary across datasets? (iii) \textbf{Consistency}: Does it converge to the correct parameter as $n\to\infty$? Beyond point estimates, \textbf{confidence sets} provide uncertainty quantification under finite sampling.

\paragraph{Causal Inference.}
Many scientific questions go beyond prediction, seeking explanations of \emph{how} variables influence one another. This requires enriching statistical models with a causal structure~\citep{pearl_causality}.
Let $\bV=\{V_1,\ldots,V_d\}$ denote \emph{endogenous variables}, quantities computed within the system. A directed graph $\mathcal{G}$ over nodes $\bV$ encodes direct causal relationships: an edge $V_i \to V_j$ indicates that $V_i$ directly causes $V_j$. A \emph{structural causal model} (SCM) is the tuple $\mathfrak{C}=(\bV,\bU,\mathbf{f},P_{\bU})$, where:

\begin{itemize}
  \item $\bU$ collects \emph{exogenous} (external) inputs representing unobserved causes or environmental randomness, $P_{\bU}$ is their joint distribution
  \item $\mathbf{f}=\{f_1,\ldots,f_d\}$ are \emph{structural assignments}, functions that deterministically compute each variable from its causes:
\end{itemize}
\begin{equation}
  V_i = f_i(\mathbf{PA}_i, U_i), \qquad i=1,\ldots,d,
\end{equation}
where $\mathbf{PA}_i\subseteq\bV$ denotes the parents of $V_i$ in $\mathcal{G}$, and $U_i\in\bU$ is its exogenous input.
SCMs enable reasoning about interventions and counterfactuals. A \emph{hard intervention} $\mathrm{do}(\bV_I=\bv_I)$ on a subset $\bV_I\subseteq\bV$ replaces the structural assignments for variables in $\bV_I$ with constants $\bv_I$, overriding their causal mechanisms. The intervened model $\mathfrak{C};\mathrm{do}(\bV_I=\bv_I)$ induces an \emph{interventional distribution} $P^{\mathfrak{C};\mathrm{do}(\bV_I=\bv_I)}_{\bV}$, which captures how the system behaves under this external manipulation.

A \emph{causal query} $q(\mathfrak{C})$ is any well-defined question about the SCM, such as ``What is the marginal distribution of $V_i$?'' or ``What is the average effect of setting $V_i=v$ on outcome $V_m$?'' Thus, a causal query is any measurable functional of the SCM, possibly involving conditioning or intervention. Central to causal inference is \emph{query identifiability}: whether $q(\mathfrak{C})$ can be uniquely determined from available observational or interventional data.

\subsection{Neural Networks as Structural Causal Models}
\label{app:nn_scm}
Returning to modern AI interpretability, we first state a standard framing of computational systems as SCMs. Let $f$ be a computational system, typically a neural network, with internal computational elements $\bV$ and input distribution $P_{\bU}$. The input distribution $P_{\bU}$ represents the \emph{behavior of interest} that we aim to explain. For instance, $P_{\bU}$ might represent arithmetic prompts to a language model, images from a particular domain, or factual questions about a specific topic.

The tuple $(f, P_{\bU})$ naturally defines an SCM $\mathfrak{C} = (\mathcal{G},\; \bV,\; \bU,\; \mathbf{f},\; P_{\bU})$, where:
\begin{itemize}
  \item \textbf{Endogenous variables} $\bV$ are the network's computational variables (e.g., hidden states, attention patterns, outputs).
  \item \textbf{Exogenous variables} $\bU$ are inputs sampled from $P_{\bU}$, representing the behavior we seek to explain.
  \item \textbf{Structural assignments} $\mathbf{f}$ are the deterministic functions defining the network's computation (layers, attention mechanisms, nonlinearities).
  \item \textbf{Causal graph} $\mathcal{G}$: the network's computation graph.
\end{itemize}

The SCM induces a unique \emph{observational distribution} over $\bV$: sampling corresponds to drawing inputs from $P_{\bU}$, executing a forward pass, and recording desired activations. Also, the SCM encodes \emph{interventional} and \emph{counterfactual} distributions based on external modifications of the inner computation.
This perspective is standard within mechanistic interpretability~\citep{olah2018the,cammarata2020thread,pmlr-v162-geiger22a,geiger2025causalabstractiontheoreticalfoundation} and is illustrated in Figure~\ref{fig:scm}. Then, a \emph{causal query} is any well-specified quantity about $\mathfrak{C} := (f, P_{\bU})$, such as ``What distribution would the network produce if we forced activation $V_i$ to value $v$?'' or ``How much does attention head $V_a$ causally contribute to correct factual recall?''
\section{Extended Related Work: Documented Failure Modes}
\label{app:related}

This appendix gives the full overview summarized in Section~\ref{sec:description}.

\medskip
\noindent\textbf{Feature Attribution.} 
Attribution methods \citep{simonyan2014deepinsideconvolutionalnetworks,sundararajan2017axiomatic} aim to highlight input features most relevant to model predictions. However, \citet{adebayo2018sanity} demonstrated that saliency maps can remain visually plausible even after model weights are randomized. Further, \citet{NEURIPS2019_bb836c01} showed that gradient-based explanations can be manipulated by adversarial perturbations, leaving predictions unchanged, while \citet{Ghorbani_Abid_Zou_2019} revealed that explanations are unstable under minor data transformations. From a theoretical standpoint, \citet{bilodeau2024impossibility} established impossibility results showing that no attribution method can simultaneously satisfy intuitive desiderata across broad model classes. 

\medskip
\noindent\textbf{Probing.} 
Probing methods train a classifier to predict a target label from internal activations. Early studies showed that both linear and structural probes could recover information with surprisingly high accuracy from randomized contextualized embeddings \citep{conneau-etal-2018-cram,hewitt-manning-2019-structural}, and syntactic probes do not generalize~\citep{hall-maudslay-cotterell-2021-syntactic}. Later, \citet{ravichander-etal-2021-probing} demonstrated that probes can extract features merely encoded (e.g., inherited from embeddings) even if unused during inference; probing asks whether a concept is encoded in an activation, not whether it is computationally relevant. Capacity-controlled probes~\citep{voita-titov-2020-information,zhu-rudzicz-2020-information,pimentel-etal-2020-information,belinkov-2022-probing} or amnesic probing~\citep{elazar-etal-2021-amnesic} attempt to mitigate such false discoveries. 

\medskip
\noindent\textbf{Sparse Autoencoders.} 
Unsupervised concept-discovery pipelines such as sparse autoencoders (SAEs) \citep{cunningham2023sparseautoencodershighlyinterpretable,yun-etal-2021-transformer,bricken2023monosemanticity,templeton2024scaling} display analogous pathologies. 
\citet{heap2025sparseautoencodersinterpretrandomly} showed that SAEs can recover apparently interpretable components even in randomly initialized transformers. Additional studies show that SAEs often fail to generalize across settings or tasks 
\citep{heindrich2025sparseautoencodersgeneralizecase,kantamneni2025sparseautoencodersusefulcase}. Also, \citet{li2025interpretabilityillusionssparseautoencoders} show SAE are sensitive to adversarial input perturbations.

\medskip
\noindent\textbf{Concept-Based Explanations.}
Concept-based methods \citep{kim2018interpretability,bau2017network} aim to identify human-interpretable concepts that align with model representations (e.g., concept activation vectors~\citep{kim2018interpretability} or network dissection~\citep{bau2017network}). These methods also face documented limitations~\citep{sinha2025comprehensivesurveyriskslimitations,aysel2025conceptbasedexplainableartificialintelligence}. Already, \citet{bolukbasi2021interpretabilityillusionbert} showed \textit{interpretability illusion} arising where activations of individual neurons in BERT may spuriously appear to encode a concept. Then, \citet{nicolson2025explainingexplainabilityrecommendationseffective} showed that concept activation scores can produce inconsistent explanations, and \citet{ramaswamy2023overlooked} show poor generalization and high sensitivity to the dataset used to infer concepts. Finally, \citet{piratla2024estimationconceptexplanationsuncertainty} further demonstrated high variance and recommended incorporating uncertainty estimation.

\medskip
\noindent\textbf{Causal Approaches.}
To address issues with prediction-based explanations, a shift toward causality-based interpretability has emerged through the use of causal mediation analysis \citep{pearl2012causal,elazar-etal-2021-amnesic,NEURIPS2020_92650b2e,meng2022locating,finlayson-etal-2021-causal,syed2023attribution,monea2024glitchmatrixlocatingdetecting,mueller2024questrightmediatorhistory}. These methods intervene on intermediate representations to quantify causal effects of components on model outputs. 
Yet recent work documents substantial fragilities and trade-offs~\citep{canby2025reliablecausalprobinginterventions}: \citet{zhang2024bestpracticesactivationpatching} showed that such approaches are sensitive to experimental design. Then, \citet{mcgrath2023hydraeffectemergentselfrepair} discovered a ``hydra effect,'' where ablating components identified as causally important fail to change behavior due to redundant causal pathways. This phenomenon, known as \textbf{overdetermination}, occurs when multiple redundant, independently sufficient causal pathways exist~\citep{schaffer2003overdetermining,sider2003s,Dyrkolbotn_2017}. Rather than isolating simple mechanisms, interventions tend to reveal overdetermined causal structures.
 
\medskip
\noindent\textbf{Mechanistic Interpretability.}
Causal approaches culminate in \textit{mechanistic interpretability} (MI), which aims to reverse-engineer networks into human-interpretable algorithms \citep{olah2020zoom}. One family of approaches (\textit{where-then-what}) first identifies circuits carrying information from inputs to outputs and then interprets their components \citep{dunefsky2024transcodersinterpretablellmfeature,davies2024cognitiverevolutioninterpretabilityexplaining,NEURIPS2023_34e1dbe9}. The second (\textit{what-then-where}) instead starts from high-level candidate algorithms and searches for causally aligned neural subspaces, using \textit{causal abstraction} metrics \citep{Geiger-etal:2022:SAIL,pmlr-v162-geiger22a,Beckers_Halpern_2019}. Despite promising demonstrations, both categories have the typical issues~\citep{sharkey2025openproblemsmechanisticinterpretability}. Subspace patching can produce \textit{interpretability illusions} by activating alternate pathways~\citep{makelov2023subspacelookingforinterpretability}, also a problem of overdetermination. Circuit explanations often fail to generalize \citep{wang2022interpretabilitywildcircuitindirect,li2025interpretationpredictbehaviorunseen} and are sensitive to minor experimental choices \citep{meloux2025mechanisticinterpretabilitystatisticalestimation}. Exhaustive studies on toy models reveal multiple incompatible explanations for both strategies, even for random networks \citep{melouxeverything}. Finally, \citet{sutter2025nonlinearrepresentationdilemmacausal} proved that, in general, existing causal abstraction methods can produce explanations for random networks.

\medskip
\noindent\textbf{Natural Language Explanations.}
Generating natural language rationales is also a possible approach \citep{marasovic-etal-2022-shot,wiegreffe-etal-2022-reframing}. However, \citet{ajwani2024llmgeneratedblackboxexplanationsadversarially} showed that LLM-generated explanations can be systematically unfaithful, confidently providing plausible-sounding justifications for predictions made for entirely different reasons. Moreover, chain-of-thought (self-)explanations are typically unfaithful to the model’s computation~\citep{lanham2023measuringfaithfulnesschainofthoughtreasoning,arcuschin2025chainofthought,turpin2023languagemodelsdontsay}. Since many plausible stories can rationalize any behavior post hoc, natural language explanations are particularly susceptible to confabulation and false positives. 



\end{document}